\documentclass{article}

\PassOptionsToPackage{numbers}{natbib}
\usepackage[final]{neurips_2025}
\usepackage[utf8]{inputenc} %
\usepackage[T1]{fontenc}    %
\usepackage{hyperref}       %
\usepackage{url}            %
\usepackage{booktabs}       %
\usepackage{amsfonts}       %
\usepackage{nicefrac}       %
\usepackage{microtype}      %
\usepackage{xcolor}         %
\usepackage{amsmath}  %
\usepackage{array}    %
\usepackage{multirow} %
\usepackage{makecell} %
\usepackage{caption}  %
\usepackage{colortbl} %

\usepackage{graphicx}

\usepackage[T1]{fontenc}

\IfFileExists{headers/config/showoverfull.config}{
	\overfullrule=1cm
}{
}

\usepackage{marginnote}

\usepackage[backgroundcolor=none,linecolor=red,textsize=footnotesize]{todonotes}

\usepackage{etoolbox}

\newbool{includeappendix}
\setbool{includeappendix}{true} %
\IfFileExists{headers/config/noappendix.config}{
	\setbool{includeappendix}{false}
}{}

\newif\ifincludeappendixx
\ifbool{includeappendix}{
	\includeappendixxtrue
}{
	\includeappendixxfalse
}

\usepackage{xr} %
\usepackage{filecontents}

\ifbool{includeappendix}{}{
	\input{appendix-labels-loader}

	\externaldocument{appendix-labels}
}

\usepackage{acro} %

\usepackage{listings}

\usepackage{textcomp}

\usepackage{xcolor}

\usepackage[scaled=0.8]{beramono}

\definecolor{ckeyword}{HTML}{7F0055}
\definecolor{ccomment}{HTML}{3F7F5F}
\definecolor{cstring}{HTML}{2A0099}

\lstdefinestyle{numbers}{
	numbers=left,
	framexleftmargin=20pt,
	numberstyle=\tiny,
	firstnumber=auto,
	numbersep=1em,
	xleftmargin=2em
}

\lstdefinestyle{layout}{
	frame=none,
	captionpos=b,
}

\lstdefinestyle{comment-style}{
	morecomment=[l]//,
	morecomment=[s]{/*}{*/},
	commentstyle={\color{ccomment}\itshape},
}

\lstdefinestyle{string-style}{
	morestring=[b]",%
	morestring=[b]',%
	stringstyle={\color{cstring}},
	showstringspaces=false,%
}

\lstdefinestyle{keyword-style}{
	keywordstyle={\ttfamily\bfseries},
	morekeywords={
		function,
		constructor,
		int,
		bool,
		return,
		returns,
		uint
	},
	morekeywords = [2]{},
	keywordstyle = [2]{\text},
	sensitive=true,
}

\lstdefinestyle{input-encoding}{
	inputencoding=utf8,
	extendedchars=true,
	literate=
	{ℝ}{$\reals$}1%
	{→}{$\rightarrow$}1%
	{α}{$\alpha$}1%
	{β}{$\beta$}1%
	{λ}{$\lambda$}1%
	{θ}{$\theta$}1%
	{ϕ}{$\phi$}1%
}

\lstdefinestyle{escaping}{
	moredelim={**[is][\color{blue}]{\%}{\%}},
	escapechar=|,
	mathescape=true
}

\lstdefinestyle{default-style}{
	basicstyle=\fontencoding{T1}\ttfamily\footnotesize,
	style=numbers,
	style=layout,
	style=comment-style,
	style=string-style,
	style=keyword-style,
	style=input-encoding,
	style=escaping,
	tabsize=2,
	upquote=true
}

\lstdefinelanguage{BASIC}{
	language=C++,
	style=default-style
}[keywords,comments,strings]%

\usepackage{hyperref,times}
\usepackage{url}
\usepackage{amsthm}
\usepackage{thm-restate}
\usepackage{mathtools}
\usepackage{algorithm}
\usepackage[noend]{algpseudocode}
\usepackage{wrapfig}
\usepackage{graphicx}
\usepackage{xspace}
\usepackage{subcaption}
\usepackage{multicol}
\usepackage{amssymb, dsfont}
\usepackage{siunitx}
\theoremstyle{plain}
\newtheorem{theorem}{Theorem}[section]

\theoremstyle{definition}
\newtheorem{definition}[theorem]{Definition}

\theoremstyle{remark}

\usepackage{amsmath,amsfonts,bm}

\def\eqref#1{equation~\ref{#1}}

\def\1{\bm{1}}

\DeclareMathAlphabet{\mathsfit}{\encodingdefault}{\sfdefault}{m}{sl}
\SetMathAlphabet{\mathsfit}{bold}{\encodingdefault}{\sfdefault}{bx}{n}

\newcommand{\R}{\mathbb{R}}

\DeclareMathOperator*{\argmin}{arg\,min}

\DeclareMathOperator{\rank}{rank}
\DeclareMathOperator{\Span}{span}

\newcommand{\dLdZ}{\frac{\partial \mathcal L}{\partial Z}}
\newcommand{\dLdW}{\frac{\partial \mathcal L}{\partial W}}
\newcommand{\dLdb}{\frac{\partial \mathcal L}{\partial b}}
\newcommand{\Qbar}{\overline{Q}}

\newcommand{\tool}{SPEAR++}

\usepackage{tikz}
\usetikzlibrary{calc,decorations,decorations.pathmorphing}
\usetikzlibrary{positioning,fit,arrows}
\usetikzlibrary{decorations.markings}
\usetikzlibrary{shapes,shapes.geometric}
\usetikzlibrary{shadows,patterns,snakes}
\usetikzlibrary{backgrounds,decorations.pathreplacing,automata}
\usetikzlibrary{intersections}
\usetikzlibrary{angles,quotes}
\usetikzlibrary{plotmarks}
\usetikzlibrary{patterns}
\usetikzlibrary{arrows.meta}
\usetikzlibrary{shapes.misc}
\usetikzlibrary{chains}

\usepackage[capitalize]{cleveref}

\makeatletter
\renewcommand\theHALG@line{\thealgorithm.\arabic{ALG@line}}
\makeatother

\newcommand{\appref}[1]{%
	\ifbool{includeappendix}{\cref{#1}}{the appendix}%
}
\newcommand{\Appref}[1]{%
	\ifbool{includeappendix}{\cref{#1}}{The appendix}%
}

\crefname{equation}{Eq.}{Eq.}
\creflabelformat{equation}{#2#1#3}
\crefname{assumption}{Asm.}{Asm.}
\creflabelformat{assumption}{#2#1#3}
\crefname{appendix}{App.}{App.}
\crefname{section}{Sec.}{Sec.}
\crefname{theorem}{Thm.}{Thm.}
\crefname{algorithm}{Alg.}{Alg.}
\crefname{corollary}{Cor.}{Cor.}
\crefname{table}{Tab.}{Tab.}

\title{\tool: Scaling Gradient Inversion via Sparsely-Used Dictionary Learning}

\begin{document}

\author{Alexander Bakarsky$^{1}$,$\quad$ Dimitar I. Dimitrov$^{2}$,$\quad$ Maximilian Baader$^{1}$,$\quad$ Martin Vechev$^{1,2}$\hfil\\
	\hspace{0em}
	$^{1}$ ETH Zurich
	\hspace{2em}
	$^{2}$ INSAIT, Sofia University "St. Kliment Ohridski"
	\hfil\\
	\texttt{\{abakarsky, mbaader, martin.vechev\}@ethz.ch}  $^{1}$\hfil\\
	\texttt{\{dimitar.iliev.dimitrov\}@insait.ai} $^{2}$\hfil\\
}

\maketitle

\vspace{-6mm}
\begin{abstract}
Federated Learning has seen an increased deployment in real-world scenarios recently, as it enables the distributed training of machine learning models without explicit data sharing between individual clients. Yet, the introduction of the so-called gradient inversion attacks has fundamentally challenged its privacy-preserving properties. Unfortunately, as these attacks mostly rely on direct data optimization without any formal guarantees, the vulnerability of real-world systems remains in dispute and requires tedious testing for each new federated deployment. To overcome these issues, recently the SPEAR attack was introduced, which is based on a theoretical analysis of the gradients of linear layers with ReLU activations. While SPEAR is an important theoretical breakthrough, the attack's practicality was severely limited by its exponential runtime in the batch size $b$. In this work, we fill this gap by applying State-of-the-Art techniques from Sparsely-Used Dictionary Learning to make the problem of gradient inversion on linear layers with ReLU activations tractable.
Our experiments demonstrate that our new attack, SPEAR++, retains all desirable properties of SPEAR, such as robustness to DP noise and FedAvg aggregation, while being applicable to 10x bigger batch sizes.

\end{abstract}

\section{Introduction}

\paragraph{Federated Learning}

Conventional machine learning techniques require ever-increasing datasets to be collected and stored in a centralized location for training, a practice that is often not viable due to institutional data-sharing policies or governmental privacy regulations such as the General Data Protection Regulation (GDPR) and the California Consumer Privacy Act (CCPA). Federated Learning~\citep{DBLP:conf/aistats/McMahanMRHA17} addresses this fundamental limitation by decoupling the model training from the need for direct access to raw data.

Instead, in a typical federated protocol, a coordinating server distributes a global model to a set of clients. Each client computes an update to the model based exclusively on its local, private dataset. These updates, rather than the data itself, are then communicated back to the server, which aggregates them into a global model that can be shared with the clients in the next communication round.

\paragraph{Gradient Inversion Attacks} Unfortunately, recent work~\citep{DBLP:conf/nips/ZhuLH19} has demonstrated that while federated clients do not explicitly share their data with the server, an honest-but-curious server can recover the clients' private data from the shared updates through the so-called Gradient Inversion Attacks (GIAs).

Currently, it is poorly understood when such attacks pose a realistic threat. This is so, as most SotA attacks rely on direct client data optimization~\citep{DBLP:conf/nips/GeipingBD020}, which provides no guarantees, is hard to theoretically analyze, and can only approximately recover the client data.

Recently, \citet{DBLP:conf/nips/DimitrovBMV24} demonstrated that exact GIAs are possible in the special case of linear layers with ReLU activations, providing a promising pathway toward theoretical analysis of gradient leakage in this setting. Yet, the practicality of the introduced method, SPEAR, remains questionable due to its exponential runtime in the batch size $b$.

\paragraph{This Work: Gradient Inversion via Sparsely-Used Dictionary Learning}
Sparsely-Used Dictionary Learning, or simply Dictionary Learning, is a classical computer science problem where one tries to find the best possible representation of a given dataset in terms of sparse linear combinations of unknown dictionary elements. While ~\citet{DBLP:conf/nips/DimitrovBMV24} already points out the connection between this problem and problem of GIAs on ReLU-activated weights, the authors stop short of exploring the full potential of Dictionary Learning in this setting. In this work, we fill this gap by exploring the applications of SotA Dictionary Learning techniques to gradient inversion and demonstrate that to a large extent they alleviate the scalability issues demonstrated by SPEAR. This in turn exposes the fundamental vulnerability of gradient updates in this setting.

\paragraph{Main Contributions:}
\begin{itemize}
	\vspace{-3.5mm}
	\setlength\itemsep{0.15em}
	\item Exploration of the connection between Dictionary Learning techniques and Gradient Inversion methods like SPEAR that take advantage of the gradient sparsity induced by ReLU.
	\item Extensive experimental evaluation of Dictionary Learning methods within the framework of SPEAR, showing they scale much more favorably and thus pose a much greater risk to practical FL deployments.
	\item Experiments on FedAvg updates and gradient defended with DP noise, showing comparable robustness between the Dictionary Learning methods and SPEAR, while allowing for enhanced scalability of the attack.
\end{itemize}

\section{Prior Work}
Gradient inversion attacks fall into two categories --- \emph{malicious} ones, where the adversary tampers with the models sent to the clients \citep{DBLP:conf/eurosp/BoenischDSSSP23, DBLP:conf/iclr/FowlGCGG22}, and \emph{honest-but-curious} ones \cite{DBLP:conf/nips/ZhuLH19, DBLP:conf/nips/GeipingBD020, DBLP:journals/corr/abs-2001-02610, DBLP:conf/nips/DimitrovBMV24, DBLP:conf/nips/PetrovDBMV24, DBLP:conf/iclr/DrenchevaPBDV25, DBLP:conf/icml/KariyappaGM0SQL23}, where the private data is reconstructed only based on observed gradients without any modifications to the federated protocol. In this work, we will focus on the latter. First GIAs in the honest setting focused on directly optimizing for the client inputs by minimizing the distance between the received and simulated gradients over dummy data~\citep{DBLP:conf/nips/ZhuLH19, DBLP:conf/nips/GeipingBD020, DBLP:journals/corr/abs-2001-02610}, achieving approximate reconstructions. 

More recently, exact gradient inversion attacks have been introduced \citep{DBLP:conf/nips/DimitrovBMV24, DBLP:conf/nips/PetrovDBMV24, DBLP:conf/iclr/DrenchevaPBDV25} that exploit the low-rank structure of the gradients of linear layers to exactly recover individual inputs from larger batches of data.  In particular, \citet{DBLP:conf/nips/PetrovDBMV24} and  \citet{DBLP:conf/iclr/DrenchevaPBDV25} leverage the low-rank to efficiently filter out incorrect text subsequences or subgraphs from a finite set of possibilities. However, this approach is limited to discrete data modalities. In contrast, \citet{DBLP:conf/nips/DimitrovBMV24} introduced SPEAR, which is domain agnostic but hindered by exponential time complexity with respect to the client batch size. In this work, we overcome this limitation by using techniques from Dictionary Learning \citep{DBLP:journals/jmlr/SpielmanWW12, DBLP:journals/tit/SunQW17, DBLP:journals/corr/abs-2001-06970} achieving the first scalable and exact domain-agnostic GIA for non-discrete modalities. We lean on the insights of \citet{DBLP:journals/tit/SunQW17} that the problem is amenable to efficient non-convex optimization via loss relaxation and leverage first-order optimization procedures to recover the sparse columns of the matrix $\dLdZ$. For losses that yield approximate solutions, we "round" them to the true ones by sampling similarly to SPEAR, but guided by the inexact solution to ensure high success rate after just a few samples.

\section{Background}
In this section, we introduce the problem setting as well as the algorithms we build upon. 

\subsection{Threat model} We consider a fully-connected linear layer $Z = \textbf{W}X + (\textbf{b}| \dots |\textbf{b})$ with parameters $\textbf{W}\in \mathbb{R}^{m\times n}$ and $\textbf{b}\in \mathbb{R}^{m}$ anywhere in the client's neural network model, activated by $Y = \text{ReLU}(Z)$. The attacker is a participant in the federated learning protocol, and hence the layer parameters $\textbf{W}$ and $\textbf{b}$ are known. Most importantly, the attacker has access to the gradients $\dLdW$ and $\dLdb$ of other participants and aims to restore the input $X$ which generated them. If the linear layer is the first layer of the network this corresponds to obtaining the client data. Otherwise, our attack recovers the intermediate features of the client data which can be used with an approximate attack (see Section 6.4 in \citet{DBLP:conf/nips/DimitrovBMV24}). We assume that each of the $b$ input datum has been vectorized to a column vector $X_i$ and the batch is formed as $X = (X_1| \dots | X_b)$.

\subsection{Exact Gradient Inversion for Batches} Exact gradient inversion in the hones-but-curious setting for batches of size $\geq 1$ was achieved by exploiting two key properties of the gradients of the network. Namely, the gradient can be explicitly expressed with respect to the client data through a low-rank decomposition:

\begin{theorem}[\citet{DBLP:conf/nips/DimitrovBMV24}]
    \label{th:low-rank}
    The network's gradient w.r.t. the weights $\textbf{W}$ can be represented as the matrix product:
    \begin{equation}
        \dLdW = \dLdZ X^\top
    \end{equation}

\end{theorem}

Since neither $X$ nor $\dLdZ$ are known a priori, the algorithm starts with an arbitrary low-rank decomposition based on SVD: $\dLdW = L R$. Then, the problem of finding the inputs is reformulated as determining the unique disaggregation matrix restoring the desired decomposition:

\begin{theorem}[{\citet{DBLP:conf/nips/DimitrovBMV24}}]
    \label{th:disaggregation}
    If the gradient $\dLdZ$ and the input matrix $X$ are of full rank and $b \leq n, m$, then there exists a unique matrix $Q \in \R^{b \times b}$ of full rank s.t. $\dLdZ = LQ$ and $X^\top = Q^{-1} R$.
\end{theorem}

The gradient of the activation $\dLdZ$ has a fraction of zero entries equal to $\frac{1}{2}$, induced by ReLU. To find the correct matrix, the algorithm leverages this naturally arising sparsity.

\begin{theorem}[{\citet{DBLP:conf/nips/DimitrovBMV24}}]
    \label{lemma:submatrix}
    Let $A \in \R^{b -1 \times b}$ be a submatrix of $\dLdZ$ such that its i-th column is $0$ for some $i \in \{1, \dots , b\}$. Further, let $\dLdZ$, $X$, and $A$ be of full rank and $Q$ be as in \cref{th:disaggregation}. Then, there exists a full-rank submatrix $L_A \in \R^{b -1 \times b}$ of $L$ such that $\Span(q_i) = \ker(L_A)$ for the i-th column $q_i$ of $Q = (q_1| \dots |q_b)$.
\end{theorem}

Since $\dLdZ$ is not known a priori, SPEAR relies on a randomized sampling procedure to pick such a full-rank submatrix $L_A$. For each such submatrix, $\hat q_i \in \ker(L_A)$ is a potential candidate for an (unscaled) column of $Q$ if $L \hat{q}_i$ is sparse enough. Since at the end of the sampling procedure one ends up with a number of candidates larger than the batch size $b$, the algorithm applies two-stage filtering in order to select the final unscaled columns $\hat{q_i}$. A crucial part of this two-stage filtering is the greedy optimization of the sparsity matching score of the candidate solution $\Qbar$.

\begin{definition} [{\citet{DBLP:conf/nips/DimitrovBMV24}}]
    Let $\lambda_{-}$ be the number of non-positive entries in $Z$ whose corresponding entries in $\dLdZ$ are 0. Similarly, let $\lambda_{+}$ be the number of positive entries in $Z$ whose corresponding entries in $\dLdZ$ are not $0$. We call their normalized sum the \emph{sparsity matching coefficient} $\lambda$:
    \begin{equation*}
        \lambda = \frac{\lambda_{-} + \lambda_{+}}{m \cdot b}
    \end{equation*}
\end{definition}

\section{Leveraging Sparsely-Used Dictionary Learning for Gradient Leakage}

The number of submatrices SPEAR samples until it picks one satisfying the assumptions of \cref{lemma:submatrix} is exponential with respect to the batch size. Thus, a focus of this work is to substitute this procedure for an algorithm that is able to pick (unscaled) candidate columns of $Q$ more efficiently and thus scale the method to previously impossible batch sizes. Due to the ReLU-induced sparsity of $\dLdZ$, restoring $Q$ fits neatly in the framework of the Dictionary Learning problem, which assumes efficient algorithms.

\subsection{Initial Decomposition}
\label{seq:init_decomp}

In the spirit of SPEAR, our gradient leakage attack starts with an initial decomposition based on SVD:

\begin{equation*}
 \dLdZ =  \underbrace{U_{:, :b}}_{L} \underbrace{\Sigma_{:b, :b} V^\top_{:b, :}}_{R} = L R,
\end{equation*}

Where $U, \Sigma, V = \text{SVD}(\dLdW)$ is the full SVD decomposition of the observed gradient. The choice for decomposition has the following justification. SPEAR starts with $\dLdW = L' R'$ for $L' = U_{:, :b} \sqrt{\Sigma_{:b, :b}}$ and $R' = \sqrt{\Sigma_{:b, :b}} V^\top_{:b, :}$. We intend to compute $Q$ by decomposing the left factor as $L' = \dLdZ Q^{-1}$ via Dictionary Learning. A common preprocessing step to help the computation of the decomposition is to precondition the observed matrix such that it appears generated from the same sparse coefficient matrix $\dLdZ$ multiplied by an almost orthogonal dictionary. This preconditioning is carried out as $L = L' ((L')^\top L')^{-1/2}$. From here, it is straightforward to see that the processed left factor assumes the form $L = U_{:, :b}$. We can derive the corresponding formula for the right factor by compensating for the preconditioning as $R = ((L')^\top L')^{1/2} R'$.

\subsection{Disaggregation Matrix Search as Dictionary Learning}
Next, leaning on \cref{th:disaggregation}, we compute the disaggregation matrix $Q$, which restores the input by $Q^{-1} R$. Instead of searching in the kernels of random submatrices of $L$, we look for (unscaled) columns of $Q$ in the framework of Dictionary Learning. The problem is concerned with decomposing an observed matrix $L$ into a sparse coefficient matrix, in our case $\dLdZ$, multiplied by a square and invertible (complete) dictionary $Q^{-1}$, i.e., $L = \dLdZ Q^{-1} \iff LQ = \dLdZ$. Even though the problem is known to be NP-hard \citep{DBLP:journals/mp/MurtyK87}, polynomial-time algorithms which succeed with high probability have been devised \citep{DBLP:journals/tit/SunQW17a, DBLP:journals/jmlr/SpielmanWW12, DBLP:conf/iclr/BaiJS19, DBLP:conf/icml/GilboaB019, DBLP:journals/simods/LiangZFZ25, DBLP:journals/jmlr/ZhaiYL0020}. Formally, the columns of $Q$ are all local minima of:

\begin{equation*}
 \argmin_{q \not = 0} \|Lq\|_0.
\end{equation*}

However, the discrete nature of the $\|\cdot\|_0$ loss hinders effective gradient-based optimization. A line of work \citep{DBLP:journals/jmlr/SpielmanWW12, DBLP:journals/tit/SunQW17a, DBLP:conf/iclr/BaiJS19, DBLP:conf/icml/GilboaB019} focuses on finding the columns of the sparse coefficient matrix one-by-one by optimizing:

\begin{equation}
  \label{eq:argminl1}
 \argmin_{q \in \mathbb{S}^{b-1}} \varphi(Lq),
\end{equation}

where $\mathbb{S}^{b-1}$ is the $\ell_2$ hypersphere and $\varphi(\cdot)$ is a convex surrogate of $\|\cdot\|_0$. There are many options for the surrogate offering different levels of smoothness. \citet{DBLP:journals/corr/abs-2001-06970} provides an extensive comparison between many possibilities. We compare the smooth $h_\mu(x) = \mu \log \circ \cosh(x/\mu)$, $-\ell_4$, and the sparsity-promoting $\ell_1$ loss and prefer the last one for its superior experimental performance. Its non-differentiability does not pose a practical issue, as subgradient descent methods optimize it effectively. Furthermore, its local minima coincide with the minima of $\|\cdot\|_0$ (see Table II in \citep{DBLP:journals/corr/abs-2001-06970}), alleviating the need for rounding (\cref{sec:round}) as with the smooth surrogates.

To find a local minimum, we initialize a guess on the sphere uniformly at random. Then, we optimize with Riemannian Adam \citep{DBLP:conf/iclr/BecigneulG19} as implemented in the Geoopt package \citep{geoopt2020kochurov}. Another possibility to carry out the optimization is through Projected Gradient Descent~\cite{madry2017towards}. We compare the two methods in multiple settings in \cref{tab:main_results}. After the optimizer has converged to a solution, like SPEAR, we add it to a candidate pool $S$ only if it is sparse enough and $S$ does not contain it already (or its negative equivalent).

\subsection{Filtering of False Candidates}
Even though the function landscape of \cref{eq:argminl1} attains favorable properties when the sample size $m$ is much greater than the batch size $b$ \citep{DBLP:journals/tit/SunQW17, DBLP:conf/iclr/BaiJS19}, the landscape gets more hostile when the ratio between $m$ and $b$ decreases. In this case, we observe an increase in false positives, indicating the possibility of spurious local minima, which do not correspond to any column of $Q$. To deal with the false positives, we apply the two-stage filtering of SPEAR. Firstly, we select the $b$ sparsest candidates of $S$ which form a full-rank matrix $\Qbar$, and we proceed to optimize the sparsity matching score by greedily swapping columns of $\Qbar$ with directions from the rest of the solution set $S$ when the sparsity matching score increases.

We note that a successful discovery of $\Qbar$ can correspond to the true disaggregation matrix $Q$ only up to permutation and scaling of the columns. We fix the scaling using the gradient with respect to the bias $\dLdb$ as per the following result.

\begin{theorem}{\citet{DBLP:conf/nips/DimitrovBMV24}}
 For any left inverse $L^\dagger$ of $L$, we have
 \begin{equation*}
        \begin{bmatrix}
        s_1 \\
        \vdots \\
         s_n
        \end{bmatrix} = \Qbar^{-1} L^{\dagger} \dLdb
    \end{equation*}
\end{theorem}

Then we rescale as $Q = \Qbar \cdot \text{diag}(s_1, \dots, s_n)$. We also note that the permutation of the columns is inconsequential, because it only leads to a permutation of the restored data in the batch. The input is finally reconstructed as $X^\top = Q^{-1} R$.

\subsection{Rounding via Sampling}\label{sec:round}
There is a discrepancy between the local minima of \cref{eq:argminl1} when choosing $\varphi(\cdot)$ to be a smooth surrogate, such as $h_\mu$ or $-\ell_4$, versus when using $\varphi(\cdot) = \|\cdot\|_0$ directly \citep{DBLP:journals/corr/abs-2001-06970}. Hence, if we find a local minimum $\hat{q}$ of the former, we have to round it to the true sparsity-inducing direction $q^*$, usually achieved through Linear Programming (LP)~\citep{dantzig2016linear}. In practice, however, we found the LP rounding to scale poorly with problem size. Borrowing from SPEAR, we devise a scalable rounding procedure to recover the true sparse solution from the approximate one. We first compute the vector $y = L \hat{q}$. We then look at the $r$ indices of $y$ closest to $0$ in absolute value. We randomly choose a subset $\mathcal{I}$ of those with  $|\mathcal{I}| = b$ and acquire the exact solution $\hat{q} \in \ker(L_A)$, where $L_A$ is the submatrix of $L$ with rows given by the indices in $\mathcal{I}$.

\begin{algorithm}[h]
\caption{RoundingViaSampling}
\begin{algorithmic}[1]
\Require{$L$, $\hat{q}$ - an approximate solution}
\State  $y \gets L \hat{q}$
\State $\mathcal{C} \gets$ the indices of the $r$ entries of $y$ with least absolute value 
\State $\mathcal{I} \gets \text{sampleWithoutReplacement}(\mathcal{C}, b)$
\State $L_A \gets$ submatrix of $L$ with rows indexed by $\mathcal{I}$
\If{$\dim (\ker(L_A)) = 1$}
    \State \Return $q^*$ s.t. $\Span(q^*) = \ker L_A$
\Else   
    \State \Return None
\EndIf

\end{algorithmic}
\end{algorithm}

\subsection{Final algorithm}
We present the pseudocode for \tool~ in \cref{code:main}. We start with the initial low-rank decomposition of the gradient as described in \cref{seq:init_decomp}. Next, we initiate first-order optimization through the $\text{OptimOnTheSphere}$ routine, which runs either Riemannian Adam or PGD. We then round the solution via sampling if we use a smooth loss, and finally add the result $q^*$ to the candidate set $S$ only if the direction is sparse enough. At the end, after choosing the sparsest linearly independent candidates and storing them in $\mathcal{B}$, we check if $\mathcal{B}$ has sufficient rank. If not, we add the vectors corresponding to an orthonormal basis of $\Span(\mathcal{B})^\bot$. This way, we achieve partial reconstruction for batches where the algorithm isn't able to find enough linearly independent vectors. If we obtain too many candidate solutions in $\mathcal{B}$, similarly to \citet{DBLP:conf/nips/DimitrovBMV24}, we use the sparsity matching coefficient $\lambda$ to greedily select the best.

\begin{algorithm}[h]
\caption{Main Routine}
\label{code:main}
\begin{algorithmic}[1]
\Require{$\dLdW$, $\dLdb$, $\mathbf{W}$, $\mathbf{b}$}
\State $L$, $R$, $b \gets \text{InitialDecomposition}(\dLdW)$
\For{$i = 1$ to $N$}
    \State $\hat{q} \gets \text{OptimOnTheSphere}(L)$
     \State $q^* \gets \text{RoundingViaSampling}(L, \hat{q})$
     \If{$\text{Sparsity}(L q^*) \geq \tau \cdot m$ and $q^* \not \in S$}
     \State $S \gets S\cup \{q^*\}$
     
     \If{$\rank(S) = b$}
    \State $\mathcal{B} \gets \text{initFilt}(L, S)$
    \State $\lambda, X' \gets \text{GreedyFilt}(L, R, \mathbf{W}, \mathbf{b}, \dLdb, \mathcal{B}, S)$
    \If{$\lambda = 1$}
    \State \Return $X'$
    \EndIf
    \EndIf
    \EndIf
\EndFor

\State $\mathcal{B} \gets \text{initFilt}(L, S)$
\State $\mathcal{B} \gets \mathcal{B} \cup \text{orthoBasisComplement}(\mathcal{B})$
\State $\lambda, X' \gets \text{GreedyFilt}(L, R, \textbf{W}, \textbf{b}, \dLdb, \mathcal{B}, S)$

\State \Return $X'$

\end{algorithmic}
\end{algorithm}

\section{Evaluation}
\begin{figure*}
	\vspace{-4mm}
	\centering
	\resizebox{1.01\linewidth}{!}{
		\hspace{-5mm}
				\begin{tikzpicture}
					\tikzstyle{toolstyle}=[dash pattern=on 3pt off 0pt]
					\tikzstyle{ibpstyle}=[dash pattern=on 5pt off 2pt]
					
					\def \dx{-0.15 cm}
					\def \dy{-0.1 cm}
					
					\def \dl{0.1cm}
					
					\def \w{1.2cm}
					
					\node[anchor=north east] (aa) at (0, 0) {\includegraphics[width=\w]{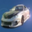}};
					\node[anchor=north west] (ab) at ($(aa.north east) + (\dx, 0 )$) {\includegraphics[width=\w]{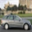}};
					\node[anchor=north west] (ac) at ($(ab.north east) + (\dx, 0 )$) {\includegraphics[width=\w]{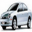}};
					\node[anchor=north west] (ad) at ($(ac.north east) + (\dx, 0 )$) {\includegraphics[width=\w]{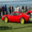}};
					\node[anchor=north west] (ae) at ($(ad.north east) + (\dx, 0 )$) {\includegraphics[width=\w]{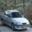}};
					\node[anchor=north west] (af) at ($(ae.north east) + (\dx, 0 )$) {\includegraphics[width=\w]{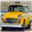}};
					\node[anchor=north west] (ag) at ($(af.north east) + (\dx, 0 )$) {\includegraphics[width=\w]{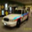}};
					\node[anchor=north west] (ah) at ($(ag.north east) + (\dx, 0 )$) {\includegraphics[width=\w]{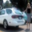}};
					\node[anchor=north west] (ai) at ($(ah.north east) + (\dx, 0 )$) {\includegraphics[width=\w]{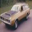}};
					\node[anchor=north west] (aj) at ($(ai.north east) + (\dx, 0 )$) {\includegraphics[width=\w]{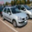}};

					\node[anchor=north east] (ca) at ($(aa.south east) + (0, -\dy )$) {\includegraphics[width=\w]{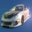}};
					\node[anchor=north west] (cb) at ($(ca.north east) + (\dx, 0 )$) {\includegraphics[width=\w]{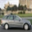}};
					\node[anchor=north west] (cc) at ($(cb.north east) + (\dx, 0 )$) {\includegraphics[width=\w]{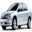}};
					\node[anchor=north west] (cd) at ($(cc.north east) + (\dx, 0 )$) {\includegraphics[width=\w]{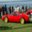}};
					\node[anchor=north west] (ce) at ($(cd.north east) + (\dx, 0 )$) {\includegraphics[width=\w]{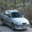}};
					\node[anchor=north west] (cf) at ($(ce.north east) + (\dx, 0 )$) {\includegraphics[width=\w]{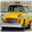}};
					\node[anchor=north west] (cg) at ($(cf.north east) + (\dx, 0 )$) {\includegraphics[width=\w]{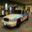}};
					\node[anchor=north west] (ch) at ($(cg.north east) + (\dx, 0 )$) {\includegraphics[width=\w]{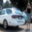}};
					\node[anchor=north west] (ci) at ($(ch.north east) + (\dx, 0 )$) {\includegraphics[width=\w]{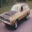}};
					\node[anchor=north west] (cj) at ($(ci.north east) + (\dx, 0 )$) {\includegraphics[width=\w]{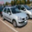}};

					\node[anchor=north east] (da) at ($(ca.south east) + (0, -\dy )$) {\includegraphics[width=\w]{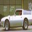}};
					\node[anchor=north west] (db) at ($(da.north east) + (\dx, 0 )$) {\includegraphics[width=\w]{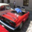}};
					\node[anchor=north west] (dc) at ($(db.north east) + (\dx, 0 )$) {\includegraphics[width=\w]{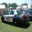}};
					\node[anchor=north west] (dd) at ($(dc.north east) + (\dx, 0 )$) {\includegraphics[width=\w]{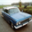}};
					\node[anchor=north west] (de) at ($(dd.north east) + (\dx, 0 )$) {\includegraphics[width=\w]{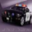}};
					\node[anchor=north west] (df) at ($(de.north east) + (\dx, 0 )$) {\includegraphics[width=\w]{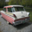}};
					\node[anchor=north west] (dg) at ($(df.north east) + (\dx, 0 )$) {\includegraphics[width=\w]{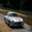}};
					\node[anchor=north west] (dh) at ($(dg.north east) + (\dx, 0 )$) {\includegraphics[width=\w]{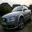}};
					\node[anchor=north west] (di) at ($(dh.north east) + (\dx, 0 )$) {\includegraphics[width=\w]{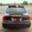}};
					\node[anchor=north west] (dj) at ($(di.north east) + (\dx, 0 )$) {\includegraphics[width=\w]{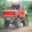}};
					
					\node[anchor=north east] (fa) at ($(da.south east) + (0, -\dy )$) {\includegraphics[width=\w]{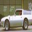}};
					\node[anchor=north west] (fb) at ($(fa.north east) + (\dx, 0 )$) {\includegraphics[width=\w]{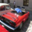}};
					\node[anchor=north west] (fc) at ($(fb.north east) + (\dx, 0 )$) {\includegraphics[width=\w]{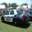}};
					\node[anchor=north west] (fd) at ($(fc.north east) + (\dx, 0 )$) {\includegraphics[width=\w]{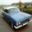}};
					\node[anchor=north west] (fe) at ($(fd.north east) + (\dx, 0 )$) {\includegraphics[width=\w]{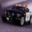}};
					\node[anchor=north west] (ff) at ($(fe.north east) + (\dx, 0 )$) {\includegraphics[width=\w]{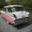}};
					\node[anchor=north west] (fg) at ($(ff.north east) + (\dx, 0 )$) {\includegraphics[width=\w]{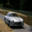}};
					\node[anchor=north west] (fh) at ($(fg.north east) + (\dx, 0 )$) {\includegraphics[width=\w]{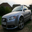}};
					\node[anchor=north west] (fi) at ($(fh.north east) + (\dx, 0 )$) {\includegraphics[width=\w]{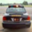}};
					\node[anchor=north west] (fj) at ($(fi.north east) + (\dx, 0 )$) {\includegraphics[width=\w]{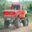}};
					
					\node[anchor=east, align=right, scale=0.8] at ($(aa.west) + (-\dl, 0 )$) {\tool~ with $\ell_1$ loss};
					\node[anchor=east, align=right, scale=0.8] at ($(ca.west) + (-\dl, 0 )$) {Original Images};
					\node[anchor=east, align=right, scale=0.8] at ($(da.west) + (-\dl, 0 )$) {\tool~ with $\ell_1$ loss};
					\node[anchor=east, align=right, scale=0.8] at ($(fa.west) + (-\dl, 0 )$) {Original Images};
					
				\end{tikzpicture}
		
	}
	\vspace{-6mm}
	\caption{All car images from a successfully reconstructed batch of size $b=210$ from CIFAR10 on network with width $m=4000$, reconstructed using \tool~ with $\ell_1$ loss and RAdam (top) compared to the ground truth (bottom). As seen, the reconstructions are indistinguishable from the original images by humans.}
	\label{fig:full_batch}
\end{figure*}
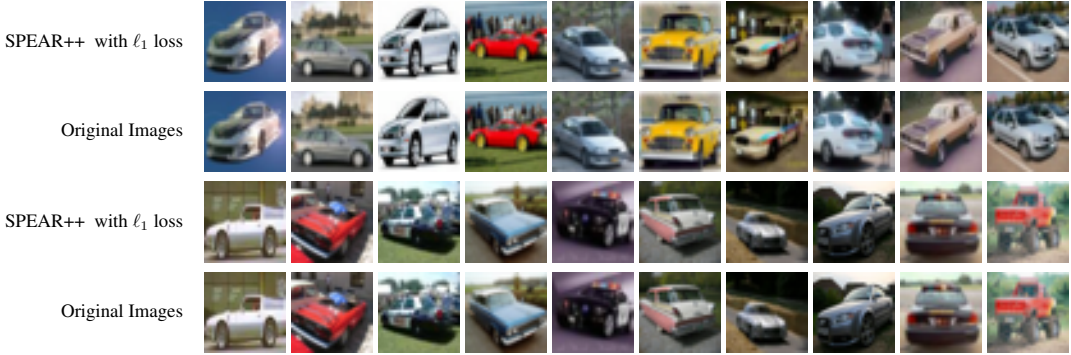

\paragraph{Setup} Unless stated otherwise, all of our experiments are conducted in the FedSGD setting using the gradients of the first layer of a fully-connected $\text{ReLU}$-activated neural network with three hidden layers, each $m$-neurons wide. The network is trained for classification on the CIFAR-10 dataset on batches of size indicated with $b$. We report the average PSNR computed across 100 reconstructed batches, as well as \tool's accuracy, which we define as the percentage of these batches with average PSNR > 90. For DP-SGD and FedAvg experiments, this threshold is lowered to 25 PSNR, as we see that for many batches in these harder settings very good reconstructions that are not pixel-perfect are often achievable. In all experiments, we use $1\mathrm{e}6$ different initializations for $q$ to allow for fair comparisons between the methods. When using $\ell_1$ for the surrogate loss $\varphi$, no rounding is applied, while for all other losses we use our rounding procedure, based on the original SPEAR paper. We optimize all initializations using Riemannian Adam on the sphere~\citep{DBLP:conf/iclr/BecigneulG19} for 500 iterations with learning rate $1\mathrm{e}\text{--}1$ which is reduced to $1\mathrm{e}\text{--}3$ and $1\mathrm{e}\text{--}5$ at the 200th and the 400th steps respectively.

\subsection{Comparing Different Dictionary Recovery Algorithms}
Inspired by \citet{DBLP:journals/corr/abs-2001-06970}, we analyze the effectiveness of different Dictionary Learning methods based on their surrogate loss $\varphi$ (\cref{tab:loss_comparison}) and optimization method (\cref{tab:main_results}) used. 

\begin{table*}[h]
\caption{Comparison between the $\ell_1$ loss (no-rounding) and the LogCosh loss function, $h_\mu$, with SPEAR-style rounding.}
\centering
\label{tab:loss_comparison}
\begin{tabular}{lcccc}
\toprule
$\varphi$        & $b$ & $m$ & PSNR & Acc (\%) \\
\midrule
\multirow{9}{*}{$h_\mu$} 
                          & 20 & 200 & 121.50 & 98 \\
                          & 25 & 200 & 106.50 & 86 \\
                          & 30 & 200 & 30.70 & 12 \\
                          \cmidrule{2-5}
                          & 65 & 1000 & 54.98 & 27 \\
                          & 100 & 1000 & 18.76 & 0 \\
                          & 150 & 1000 & 12.24 & 0 \\
                          \cmidrule{2-5}
                          & 100 & 4000 & 93.63 & 92 \\
                          & 150 & 4000 & 32.63 & 0 \\
                          & 210 & 4000 & 13.80 & 0 \\

\midrule
\multirow{9}{*}{$\ell_1$} & 20 & 200 & 120.98 &  95 \\
                          & 25 & 200 & 96.02 &  77  \\
                          & 30 & 200 & 41.70 &   25 \\
                          \cmidrule{2-5}
                          & 65  & 1000 & 125.18  &  100 \\
                          & 100 & 1000 & 69.31 &  41 \\
                          & 150 & 1000 & 10.42 &  0 \\
                          \cmidrule{2-5}
                          & 100 & 4000 & 124.24 &  100 \\
                          & 150 & 4000 & 120.86 &   100 \\
                          & 210 & 4000 & 45.34 &     7 \\
                          \midrule
\multirow{2}{*}{$-\ell_4$} & 15 & 200 & 83.02 &  62 \\
                           & 20 & 200 & 14.29 &  0  \\
                           
\bottomrule
\end{tabular}
\end{table*}

In \cref{tab:loss_comparison}, we compare $\ell_1$ to other choices of losses, including the commonly used LogCosh loss $h_\mu$, which is a smooth surrogate to the  $\ell_1$, as well as the $-\ell_4$ loss. For the  $h_\mu$, we set $\mu=300$ for the $m=200$ runs and otherwise $\mu=500$, as it worked the best in our experiments. For our rounding procedure, we sample once per reconstruction from the smallest $r=1.5b$ entries in absolute value, with the exception of the $m=200$ experiments, where we sampled from the $r=3b$ smallest entries instead. 

We observe that $\ell_1$ scales favorably with $m$, allowing reconstructions from larger batch sizes. We also observe that for smaller $m$, LogCosh is a good alternative to $\ell_1$. Importantly, both versions of \tool~ are much more scalable compared to the original SPEAR algorithm, whose runtime for $b=24$ is already prohibitively long (See Figure 4 in \citet{DBLP:conf/nips/DimitrovBMV24}). These results demonstrate a polynomial runtime relationship in $b$, unlike the exponential relationship reported in SPEAR. We further note that for $b=100$, \tool~ produces similar recovery rates to the SPEAR+Geiping combination, which, unlike \tool, relies on image priors (See Table 5 in \citet{DBLP:conf/nips/DimitrovBMV24}) at half the network width. This suggests that \tool~ is much more scalable than SPEAR, even when the latter is supplied with very strong prior information. Finally, our preliminary results on the $-\ell_4$ loss indicated even worse reconstructions than SPEAR, and thus we didn't experiment with it further.

\begin{table*}
	\centering
	\vspace{-3em}
	\caption{Comparison between reconstructions with the $\ell_1$ loss and different optimizers.}
	\label{tab:main_results}
	\begin{tabular}{lcccc}
		\toprule
		Optimizer & $b$ & $m$ & PSNR & Acc (\%) \\
		\midrule
		\multirow{9}{*}{PGD} & 20 & 200 & 41.53 &    25  \\
		& 25 & 200 & 19.01 &    0 \\
		& 30 & 200 & 14.71 &     0 \\
		\cmidrule{2-5}
		& 65  & 1000 & 93.91 &     93 \\
		& 100 & 1000 & 23.41 &   0 \\
		& 150 & 1000 & 12.06 &   0 \\
		\cmidrule{2-5}
		& 100 & 4000 & 105.61 &  100 \\
		& 150 & 4000 & 104.46 & 100 \\
		& 210 & 4000 & 88.36 &   77 \\
		\midrule
		\multirow{9}{*}{RAdam} & 20 & 200 & 120.98 &  95 \\
		& 25 & 200 & 96.02 &  77  \\
		& 30 & 200 & 41.70 &   25 \\
		\cmidrule{2-5}
		& 65  & 1000 & 125.18  &  100 \\
		& 100 & 1000 & 69.31 &  41 \\\
		& 150 & 1000 & 10.42 &  0 \\
		\cmidrule{2-5}
		& 100 & 4000 & 124.24 &  100 \\
		& 150 & 4000 & 120.86 &   100 \\
		& 210 & 4000 & 45.34 &     7 \\
		\bottomrule
	\end{tabular}
\end{table*}

Next, in  \cref{tab:main_results} we compare the reconstruction accuracy of our algorithm when optimizing the $\ell_1$ loss on the sphere using Riemannian Adam versus regular gradient descent, where we project back to the sphere at each iteration. The latter corresponds to the classical Projected Gradient Descent (PGD) algorithm~\citep{madry2017towards}. For PGD, we use the same number of steps, but a smaller initial learning rate of $1\mathrm{e}\text{--}2$. We again lower it to $1\mathrm{e}\text{--}4$ and $1\mathrm{e}\text{--}6$ at the 200th and 400th optimization step.

We generally observe Riemannian Adam to be better across the board, except in the $b=210$ experiment. While investigating this phenomenon further remains a future work item, our preliminary experiments suggest that Riemannian Adam will be able to match and exceed the PGD performance if more initializations are provided to both algorithms. 

An important observation about our experiments, regardless of the optimizer used, is that for each $m$, we practically find an upper bound on $b$, beyond which reconstruction starts failing. The ratio between the upper bound on $b$ and $m$ seems to be slowly decreasing when $m$ increases, suggesting a slightly worse than linear relationship between the upper bound and $m$. This is consistent with recent theoretical analysis of the sparse dictionary learning problem~\citep{DBLP:conf/nips/HuH23b}. 

Finally, we visualize a part of a correctly recovered batch ($b=210$) in \cref{fig:full_batch}, reaffirming the excellent quality of our reconstructions, similarly to the original SPEAR algorithm.

\subsection{Effectiveness under DP-SGD Noise}

A surprising feature of SPEAR is its effectiveness on gradients defended with DP-SGD. We consider the best-performing version of \tool, which optimizes directly $\ell_1$ with RAdam and applies no rounding, and show in \cref{tab:dp} that it matches SPEAR's performance even in the case where the noise level is similar to the median of the absolute value of the entries in the gradient for $b =20$. Our algorithm is still robust to some considerable levels of noise even on very large batches $b = 150$, considering that the median of the gradients of a $4000$-neurons-wide network is on the order of $1\mathrm{e}\text{--}5$.
\begin{table}[h]
	\centering
	\caption{Experiments on gradients protected with Differential Privacy at different noise levels $\sigma$ and clipping $C$.}
	\label{tab:dp}
	\begin{tabular}{ccccccc}
		\toprule
		$C$ & $\sigma$ & $b$ & $m$ & PSNR & Acc (\%)\\
		\midrule
		2 & $1\mathrm{e}\text{--}4$ & 20  &  200 & 31.90 & 75\\
		2 & $1\mathrm{e}\text{--}6$ & 150 & 4000 & 55.45 & 88\\
		\bottomrule 
	\end{tabular}
\end{table}

\subsection{Effectiveness under FedAvg Aggregation}

Similarly to SPEAR, we experiment with the ability of \tool~to recover data from FedAvg updates, computed with learning rate of $1\mathrm{e}\text{--}2$. We note that SPEAR's theoretical analysis and FedAvg extension (See Appendix F in \citet{DBLP:conf/nips/DimitrovBMV24}) remain valid for \tool, as well. We report the results in \cref{tab:fedavg}, where we observe that even under many local steps with unknown subsets of the client data, our attack remains effective. Further, it seems that for larger $m$, the algorithm is more robust to a large number of local gradient steps.

\begin{table}[h]
	\centering
	\caption{Experiments on FedAvg updates computed with different number of epochs $E$, and different mini-batch sizes $b_\text{mini}$.}
	\label{tab:fedavg}
	\begin{tabular}{ccccccc}
		\toprule
		$E$ & $b_\text{mini}$ & $b$ & $m$ & PSNR & Acc (\%)\\
		\midrule
		3  & 5  &  20  & 200  & 67.81 & 75\\
		15  & 30 & 150  & 4000 & 63.71 & 92 \\
		\bottomrule 
	\end{tabular}
\end{table}

\message{^^JLASTBODYPAGE \thepage^^J}
\clearpage
\subsubsection*{Acknowledgments}
This research was partially funded by the Ministry of Education and Science of Bulgaria (support for INSAIT, part of the Bulgarian National Roadmap for Research Infrastructure).

\bibliography{references}

\begin{thebibliography}{24}
\providecommand{\natexlab}[1]{#1}
\providecommand{\url}[1]{\texttt{#1}}
\expandafter\ifx\csname urlstyle\endcsname\relax
  \providecommand{\doi}[1]{doi: #1}\else
  \providecommand{\doi}{doi: \begingroup \urlstyle{rm}\Url}\fi

\bibitem[Bai et~al.(2019)Bai, Jiang, and Sun]{DBLP:conf/iclr/BaiJS19}
Yu~Bai, Qijia Jiang, and Ju~Sun.
\newblock Subgradient descent learns orthogonal dictionaries.
\newblock In \emph{7th International Conference on Learning Representations,
  {ICLR} 2019, New Orleans, LA, USA, May 6-9, 2019}. OpenReview.net, 2019.
\newblock URL \url{https://openreview.net/forum?id=HklSf3CqKm}.

\bibitem[B{\'{e}}cigneul and Ganea(2019)]{DBLP:conf/iclr/BecigneulG19}
Gary B{\'{e}}cigneul and Octavian{-}Eugen Ganea.
\newblock Riemannian adaptive optimization methods.
\newblock In \emph{7th International Conference on Learning Representations,
  {ICLR} 2019, New Orleans, LA, USA, May 6-9, 2019}. OpenReview.net, 2019.
\newblock URL \url{https://openreview.net/forum?id=r1eiqi09K7}.

\bibitem[Boenisch et~al.(2023)Boenisch, Dziedzic, Schuster, Shamsabadi,
  Shumailov, and Papernot]{DBLP:conf/eurosp/BoenischDSSSP23}
Franziska Boenisch, Adam Dziedzic, Roei Schuster, Ali~Shahin Shamsabadi, Ilia
  Shumailov, and Nicolas Papernot.
\newblock When the curious abandon honesty: Federated learning is not private.
\newblock In \emph{8th {IEEE} European Symposium on Security and Privacy,
  EuroS{\&}P 2023, Delft, Netherlands, July 3-7, 2023}, pages 175--199. {IEEE},
  2023.
\newblock \doi{10.1109/EUROSP57164.2023.00020}.
\newblock URL \url{https://doi.org/10.1109/EuroSP57164.2023.00020}.

\bibitem[Dantzig(2016)]{dantzig2016linear}
George~B Dantzig.
\newblock Linear programming and extensions.
\newblock 2016.

\bibitem[Dimitrov et~al.(2024)Dimitrov, Baader, M{\"{u}}ller, and
  Vechev]{DBLP:conf/nips/DimitrovBMV24}
Dimitar~I. Dimitrov, Maximilian Baader, Mark~Niklas M{\"{u}}ller, and Martin~T.
  Vechev.
\newblock {SPEAR:} exact gradient inversion of batches in federated learning.
\newblock In Amir Globersons, Lester Mackey, Danielle Belgrave, Angela Fan,
  Ulrich Paquet, Jakub~M. Tomczak, and Cheng Zhang, editors, \emph{Advances in
  Neural Information Processing Systems 38: Annual Conference on Neural
  Information Processing Systems 2024, NeurIPS 2024, Vancouver, BC, Canada,
  December 10 - 15, 2024}, 2024.
\newblock URL
  \url{http://papers.nips.cc/paper\_files/paper/2024/hash/c13cd7feab4beb1a27981e19e2455916-Abstract-Conference.html}.

\bibitem[Drencheva et~al.(2025)Drencheva, Petrov, Baader, Dimitrov, and
  Vechev]{DBLP:conf/iclr/DrenchevaPBDV25}
Maria Drencheva, Ivo Petrov, Maximilian Baader, Dimitar~Iliev Dimitrov, and
  Martin~T. Vechev.
\newblock {GRAIN:} exact graph reconstruction from gradients.
\newblock In \emph{The Thirteenth International Conference on Learning
  Representations, {ICLR} 2025, Singapore, April 24-28, 2025}. OpenReview.net,
  2025.
\newblock URL \url{https://openreview.net/forum?id=7bAjVh3CG3}.

\bibitem[Fowl et~al.(2022)Fowl, Geiping, Czaja, Goldblum, and
  Goldstein]{DBLP:conf/iclr/FowlGCGG22}
Liam~H. Fowl, Jonas Geiping, Wojciech Czaja, Micah Goldblum, and Tom Goldstein.
\newblock Robbing the fed: Directly obtaining private data in federated
  learning with modified models.
\newblock In \emph{The Tenth International Conference on Learning
  Representations, {ICLR} 2022, Virtual Event, April 25-29, 2022}.
  OpenReview.net, 2022.
\newblock URL \url{https://openreview.net/forum?id=fwzUgo0FM9v}.

\bibitem[Geiping et~al.(2020)Geiping, Bauermeister, Dr{\"{o}}ge, and
  Moeller]{DBLP:conf/nips/GeipingBD020}
Jonas Geiping, Hartmut Bauermeister, Hannah Dr{\"{o}}ge, and Michael Moeller.
\newblock Inverting gradients - how easy is it to break privacy in federated
  learning?
\newblock In Hugo Larochelle, Marc'Aurelio Ranzato, Raia Hadsell,
  Maria{-}Florina Balcan, and Hsuan{-}Tien Lin, editors, \emph{Advances in
  Neural Information Processing Systems 33: Annual Conference on Neural
  Information Processing Systems 2020, NeurIPS 2020, December 6-12, 2020,
  virtual}, 2020.
\newblock URL
  \url{https://proceedings.neurips.cc/paper/2020/hash/c4ede56bbd98819ae6112b20ac6bf145-Abstract.html}.

\bibitem[Gilboa et~al.(2019)Gilboa, Buchanan, and
  Wright]{DBLP:conf/icml/GilboaB019}
Dar Gilboa, Sam Buchanan, and John Wright.
\newblock Efficient dictionary learning with gradient descent.
\newblock In Kamalika Chaudhuri and Ruslan Salakhutdinov, editors,
  \emph{Proceedings of the 36th International Conference on Machine Learning,
  {ICML} 2019, 9-15 June 2019, Long Beach, California, {USA}}, volume~97 of
  \emph{Proceedings of Machine Learning Research}, pages 2252--2259. {PMLR},
  2019.
\newblock URL \url{http://proceedings.mlr.press/v97/gilboa19a.html}.

\bibitem[Hu and Huang(2023)]{DBLP:conf/nips/HuH23b}
Jingzhou Hu and Kejun Huang.
\newblock Global identifiability of l1-based dictionary learning via matrix
  volume optimization.
\newblock \emph{Advances in Neural Information Processing Systems},
  36:\penalty0 36165--36186, 2023.

\bibitem[Kariyappa et~al.(2023)Kariyappa, Guo, Maeng, Xiong, Suh, Qureshi, and
  Lee]{DBLP:conf/icml/KariyappaGM0SQL23}
Sanjay Kariyappa, Chuan Guo, Kiwan Maeng, Wenjie Xiong, G.~Edward Suh,
  Moinuddin~K. Qureshi, and Hsien{-}Hsin~S. Lee.
\newblock Cocktail party attack: Breaking aggregation-based privacy in
  federated learning using independent component analysis.
\newblock In Andreas Krause, Emma Brunskill, Kyunghyun Cho, Barbara Engelhardt,
  Sivan Sabato, and Jonathan Scarlett, editors, \emph{International Conference
  on Machine Learning, {ICML} 2023, 23-29 July 2023, Honolulu, Hawaii, {USA}},
  volume 202 of \emph{Proceedings of Machine Learning Research}, pages
  15884--15899. {PMLR}, 2023.
\newblock URL \url{https://proceedings.mlr.press/v202/kariyappa23a.html}.

\bibitem[Kochurov et~al.(2020)Kochurov, Karimov, and
  Kozlukov]{geoopt2020kochurov}
Max Kochurov, Rasul Karimov, and Serge Kozlukov.
\newblock Geoopt: Riemannian optimization in pytorch, 2020.

\bibitem[Liang et~al.(2025)Liang, Zhang, Fattahi, and
  Zhang]{DBLP:journals/simods/LiangZFZ25}
Geyu Liang, Gavin Zhang, Salar Fattahi, and Richard~Y. Zhang.
\newblock Simple alternating minimization provably solves complete dictionary
  learning.
\newblock \emph{{SIAM} J. Math. Data Sci.}, 7\penalty0 (3):\penalty0 855--883,
  2025.
\newblock \doi{10.1137/23M1568120}.
\newblock URL \url{https://doi.org/10.1137/23m1568120}.

\bibitem[Madry et~al.(2017)Madry, Makelov, Schmidt, Tsipras, and
  Vladu]{madry2017towards}
Aleksander Madry, Aleksandar Makelov, Ludwig Schmidt, Dimitris Tsipras, and
  Adrian Vladu.
\newblock Towards deep learning models resistant to adversarial attacks.
\newblock \emph{arXiv preprint arXiv:1706.06083}, 2017.

\bibitem[McMahan et~al.(2017)McMahan, Moore, Ramage, Hampson, and
  y~Arcas]{DBLP:conf/aistats/McMahanMRHA17}
Brendan McMahan, Eider Moore, Daniel Ramage, Seth Hampson, and
  Blaise~Ag{\"{u}}era y~Arcas.
\newblock Communication-efficient learning of deep networks from decentralized
  data.
\newblock In Aarti Singh and Xiaojin~(Jerry) Zhu, editors, \emph{Proceedings of
  the 20th International Conference on Artificial Intelligence and Statistics,
  {AISTATS} 2017, 20-22 April 2017, Fort Lauderdale, FL, {USA}}, volume~54 of
  \emph{Proceedings of Machine Learning Research}, pages 1273--1282. {PMLR},
  2017.
\newblock URL \url{http://proceedings.mlr.press/v54/mcmahan17a.html}.

\bibitem[Murty and Kabadi(1987)]{DBLP:journals/mp/MurtyK87}
Katta~G. Murty and Santosh~N. Kabadi.
\newblock Some np-complete problems in quadratic and nonlinear programming.
\newblock \emph{Math. Program.}, 39\penalty0 (2):\penalty0 117--129, 1987.
\newblock \doi{10.1007/BF02592948}.
\newblock URL \url{https://doi.org/10.1007/BF02592948}.

\bibitem[Petrov et~al.(2024)Petrov, Dimitrov, Baader, M{\"{u}}ller, and
  Vechev]{DBLP:conf/nips/PetrovDBMV24}
Ivo Petrov, Dimitar~I. Dimitrov, Maximilian Baader, Mark~Niklas M{\"{u}}ller,
  and Martin~T. Vechev.
\newblock {DAGER:} exact gradient inversion for large language models.
\newblock In Amir Globersons, Lester Mackey, Danielle Belgrave, Angela Fan,
  Ulrich Paquet, Jakub~M. Tomczak, and Cheng Zhang, editors, \emph{Advances in
  Neural Information Processing Systems 38: Annual Conference on Neural
  Information Processing Systems 2024, NeurIPS 2024, Vancouver, BC, Canada,
  December 10 - 15, 2024}, 2024.
\newblock URL
  \url{http://papers.nips.cc/paper\_files/paper/2024/hash/9ff1577a1f8308df1ccea6b4f64a103f-Abstract-Conference.html}.

\bibitem[Qu et~al.(2020)Qu, Zhu, Li, Tsakiris, Wright, and
  Vidal]{DBLP:journals/corr/abs-2001-06970}
Qing Qu, Zhihui Zhu, Xiao Li, Manolis~C. Tsakiris, John Wright, and Ren{\'{e}}
  Vidal.
\newblock Finding the sparsest vectors in a subspace: Theory, algorithms, and
  applications.
\newblock \emph{CoRR}, abs/2001.06970, 2020.
\newblock URL \url{https://arxiv.org/abs/2001.06970}.

\bibitem[Spielman et~al.(2012)Spielman, Wang, and
  Wright]{DBLP:journals/jmlr/SpielmanWW12}
Daniel~A. Spielman, Huan Wang, and John Wright.
\newblock Exact recovery of sparsely-used dictionaries.
\newblock In Shie Mannor, Nathan Srebro, and Robert~C. Williamson, editors,
  \emph{{COLT} 2012 - The 25th Annual Conference on Learning Theory, June
  25-27, 2012, Edinburgh, Scotland}, volume~23 of \emph{{JMLR} Proceedings},
  pages 37.1--37.18. JMLR.org, 2012.
\newblock URL \url{http://proceedings.mlr.press/v23/spielman12/spielman12.pdf}.

\bibitem[Sun et~al.(2017{\natexlab{a}})Sun, Qu, and
  Wright]{DBLP:journals/tit/SunQW17}
Ju~Sun, Qing Qu, and John Wright.
\newblock Complete dictionary recovery over the sphere {I:} overview and the
  geometric picture.
\newblock \emph{{IEEE} Trans. Inf. Theory}, 63\penalty0 (2):\penalty0 853--884,
  2017{\natexlab{a}}.
\newblock \doi{10.1109/TIT.2016.2632162}.
\newblock URL \url{https://doi.org/10.1109/TIT.2016.2632162}.

\bibitem[Sun et~al.(2017{\natexlab{b}})Sun, Qu, and
  Wright]{DBLP:journals/tit/SunQW17a}
Ju~Sun, Qing Qu, and John Wright.
\newblock Complete dictionary recovery over the sphere {II:} recovery by
  riemannian trust-region method.
\newblock \emph{{IEEE} Trans. Inf. Theory}, 63\penalty0 (2):\penalty0 885--914,
  2017{\natexlab{b}}.
\newblock \doi{10.1109/TIT.2016.2632149}.
\newblock URL \url{https://doi.org/10.1109/TIT.2016.2632149}.

\bibitem[Zhai et~al.(2020)Zhai, Yang, Liao, Wright, and
  Ma]{DBLP:journals/jmlr/ZhaiYL0020}
Yuexiang Zhai, Zitong Yang, Zhenyu Liao, John Wright, and Yi~Ma.
\newblock Complete dictionary learning via l4-norm maximization over the
  orthogonal group.
\newblock \emph{J. Mach. Learn. Res.}, 21:\penalty0 165:1--165:68, 2020.
\newblock URL \url{https://jmlr.org/papers/v21/19-755.html}.

\bibitem[Zhao et~al.(2020)Zhao, Mopuri, and
  Bilen]{DBLP:journals/corr/abs-2001-02610}
Bo~Zhao, Konda~Reddy Mopuri, and Hakan Bilen.
\newblock idlg: Improved deep leakage from gradients.
\newblock \emph{CoRR}, abs/2001.02610, 2020.
\newblock URL \url{http://arxiv.org/abs/2001.02610}.

\bibitem[Zhu et~al.(2019)Zhu, Liu, and Han]{DBLP:conf/nips/ZhuLH19}
Ligeng Zhu, Zhijian Liu, and Song Han.
\newblock Deep leakage from gradients.
\newblock In Hanna~M. Wallach, Hugo Larochelle, Alina Beygelzimer, Florence
  d'Alch{\'{e}}{-}Buc, Emily~B. Fox, and Roman Garnett, editors, \emph{Advances
  in Neural Information Processing Systems 32: Annual Conference on Neural
  Information Processing Systems 2019, NeurIPS 2019, December 8-14, 2019,
  Vancouver, BC, Canada}, pages 14747--14756, 2019.
\newblock URL
  \url{https://proceedings.neurips.cc/paper/2019/hash/60a6c4002cc7b29142def8871531281a-Abstract.html}.

\end{thebibliography}
\bibliographystyle{plainnat}

\message{^^JLASTREFERENCESPAGE \thepage^^J}

\ifincludeappendixx
	\clearpage
	\appendix
	
\section{Deferred Algorithms}

\begin{algorithm}[h]
    \caption{initFilt, adapted from \protect\citet{DBLP:conf/nips/DimitrovBMV24}}
    
    \begin{algorithmic}[1]
    \Require{$L$, $S$}
       
    \State $\mathcal{B} \gets \emptyset$
    \State $\overline{\mathcal{B}} \gets \emptyset$

    \While{$\dim \Span(\mathcal{B}) < b$ and $S \setminus (\mathcal{B} \cup \overline{\mathcal{B}}) \not = \emptyset$}
        \State Select the vector $q$ from $S \setminus (\mathcal{B} \cup \overline{\mathcal{B}})$ inducing the sparsest $Lq$
        \State $\mathcal{B} \gets \mathcal{B} \cup \{q\}$
        \If{ $\dim(\Span(\mathcal{B})) \not = |\mathcal{B}|$}
            \State $\mathcal{B} \gets \mathcal{B} \setminus \{q\}$
            \State $\overline{\mathcal{B}} \gets \overline{\mathcal{B}}  \cup \{q\}$
        \EndIf
    \EndWhile

    \State \Return $\mathcal{B}$
    \end{algorithmic}
\end{algorithm}

\begin{algorithm}[h]
\caption{GreedyFilt, adapted from \protect\citet{DBLP:conf/nips/DimitrovBMV24}}
\begin{algorithmic}[1]
\Require{$L$, $R$, $\mathbf{W}$, $\mathbf{b}$, $\dLdb$, $\mathcal{B}$, $S$}

    \State $Q$ initialized with columns from $\mathcal{B}$
    \State $Q \gets \text{fixScale}(Q, L, \dLdb)$
    \State $\lambda \gets \text{computeScore}(L, R, Q, \mathbf{W}, \mathbf{b})$
    \If{$\lambda = 1$}
        \State \Return $\lambda$, $(Q^{-1} R)^\top$
    \EndIf

    \For{$s \in S \setminus \mathcal{B}$}
        \For {$j \in \{1, \dots, b\}}$
            \State $Q' \gets Q$
            \State $(Q')_{:, j} \gets s$
            \If {$\text{rank}(Q') < b$}
                \State continue
            \EndIf
            \State $Q' \gets \text{fixScale}(Q', L, \dLdb)$
            \State $\lambda' \gets \text{computeScore}(L, R, Q', \mathbf{W}, \mathbf{b})$
            \If{$\lambda' > \lambda$}
                \State $\lambda \gets \lambda'$
                \State $Q \gets Q'$
            \EndIf

        \EndFor
    \EndFor

    \State \Return $\lambda, (Q^{-1}R)^{\top}$
    
\end{algorithmic}
\end{algorithm}

\begin{algorithm}[h]
\caption{fixScale, adapted from \protect\citet{DBLP:conf/nips/DimitrovBMV24}}
\begin{algorithmic}[1]
\Require{$\hat Q$, $L$, $\dLdb$}
\State $d \gets \hat{Q}^{-1} L^\dagger \dLdb$
\State $Q \gets \hat{Q} \cdot \text{diag}(d)$
\State \Return $Q$
\end{algorithmic}
\end{algorithm}

\begin{algorithm}[h]
\caption{computeScore, adapted from \protect\citet{DBLP:conf/nips/DimitrovBMV24}}
\begin{algorithmic}[1]
\Require{$L$, $R$, $Q$, $\mathbf{W}$, $\mathbf{b}$}
\State $Z' \gets \mathbf{W}(Q^{-1} R)^\top + (\mathbf{b}| \dots |\mathbf{b})$
\State $\dLdZ' \gets LQ$
\State $\lambda_{-} \gets \sum_{i, j} \mathds{1}[Z_{i,j}' \leq 0] \cdot \mathds{1}[\Big(\dLdZ'\Big)_{i,j} = 0]$
\State $\lambda_{+} \gets \sum_{i, j} \mathds{1}[Z_{i,j}' > 0] \cdot \mathds{1}[\Big( \dLdZ' \Big)_{i, j} \not = 0]$
\State $\lambda \gets \frac{\lambda_{-} + \lambda_{+}}{m \cdot b}$
\State \Return $\lambda$
\end{algorithmic}
\end{algorithm}

\fi

\end{document}